\documentclass{article}

\usepackage[utf8]{inputenc} % allow utf-8 input
\usepackage[T1]{fontenc}    % use 8-bit T1 fonts
\usepackage{hyperref}       % hyperlinks
\usepackage{url}            % simple URL typesetting
\usepackage{booktabs}       % professional-quality tables
\usepackage{amsfonts}       % blackboard math symbols
\usepackage{nicefrac}       % compact symbols for 1/2, etc.
\usepackage{microtype}      % microtypography
\usepackage[pdftex]{graphicx}

\title{A Game of Dice: Machine Learning and the Question Concerning Art}

\author{
  Paul Todorov \\
  \texttt{paul.todorov@hec.edu} \\
}
\date{\vspace{-5ex}}
\begin{document}

\maketitle

\begin{abstract}
By this art you may contemplate the variation of the 26 letters.\\

We review some practical and philosophical questions raised by the use of machine learning in creative practice. Beyond the obvious problems regarding plagiarism and authorship, we argue that the novelty in AI Art relies mostly on a narrow machine learning contribution : manifold approximation. Nevertheless, this contribution creates a radical shift in the way we have to consider this movement. Is this omnipotent tool a blessing or a curse for the artists?

\end{abstract}

\section{Plagiarism and Authorship}
Plagiarism is not a very well defined term in the field of art. Appropriation (inspiration, collage, tribute) has always been at the heart of the creative process, yet, the use of machine learning brings additional difficulties. It may seem paradoxical or ironical since there is also hope that deep learning could actually help fight against art forgery [1]. Most of the AI art practitioners use algorithms derived from generative adversarial networks [2]. In this type of work, the artist has to (1) define a training set, (2) choose the algorithm (GAN variant), choose the hyper parameters and the optimization method, (3) select samples from a latent space, (4) generate outputs and possibly post-process them.\\

Why is plagiarism a tricky issue with AI Art? We can trivially show that every painting is a geometric transformation of another painting. Consequently, the problem for AI art isn't that they can be obtained as a geometric transformation of other paintings, but that it is precisely the way they are created. By doing so, it confronts us to a threshold phenomenon. Everyone would agree that outputs from an algorithm that is  able to perfectly memorize and reproduce a training set, or that do not deteriorate at all a copyrighted input, would be plagiarism. But what if we imagine the following simple example: training an autoencoder with one hidden layer large enough to learn the identity function on a set of copyrighted paintings. Then, at different points in time during training we apply the autoencoder to a sample from the training set. We end up with a full collection of paintings, closer and closer to the original one.\\

Which of those are genuine pieces of AI Art, and which of those are plagiarism? This is a very practical question since this kind of process is central in most of today's AI Art. In addition, because of the open source paradigm in machine learning, sophisticated methods are becoming increasingly available and ready for use. Hence authorship becomes problematic as well. Can we still consider machine learning as a tool when this tool is so powerful that we just need to press a button? It is unclear whether we should credit the algorithm, the one who implemented the pipeline or the one who used it. The recent controversy surrounding the work from the French collective Obvious [3] shows that those questions remain unanswered.  Those blatant problems might actually only be the consequences of a more radical problem introduced by the use of machine learning for art creation. What if man becomes a tool as well?

\section{Novelty in AI Art}

Using GANs for art creation is indubitably groundbreaking. Yet, it is groundbreaking mostly regarding the way the generator is obtained. The core principle: sampling from a latent space and using a generator to obtain pieces of art is a process that has already been used in the past. A most striking example is the book by Raymond Queneau, \textit{Cent Mille Milliards de Poèmes} [4] published in 1961.The book is composed of ten sonnets printed on separate cards. Each line is printed on a separate strip. Since all ten sonnets have the same rhyme scheme and rhyme sounds, any lines from a sonnet can be combined with any from the nine others. By turning the strips, the reader can then create up to $10^{14}$ poems. This is actually very similar to what AI art currently does.\\

Let $L1,…,L14$ be the lists that index the 10 different verses associated with the different rows of the final sonnet. We can define $G(x1,…,x14) = [L1[x1],...,L14[x14]]$ which is equivalent to turning the strips as if they were pages, choosing the $x1$ th page  for the first row, etc. This can be considered the intuitive use of the book. A poem can then be obtained by doing the following: (i) Randomly sample 14 numbers between 1 and 10: $(a1,...,a14)$, (ii) Use a generator $G$, so as to obtain the output $p = G(a1,...,a14)$\\

Raymond Queneau has provided the readers with a generator, allowing them to generate poems themselves by sampling from a finite countable latent space. It is interesting to see that this primitive work differentiates itself from the current trend, where only the outputs are sold by the artists. One can say that the main novelty that machine learning brings compared to this work is our ability to approximate interesting manifolds (of “existing images” or “understandable text”), whereas Queneau had to manually design all the process. Regarding plagiarism and authorship, this piece of art has a very informative anecdote. In 1997 someone published \textit{Cent Mille Milliards de Poèmes} online, setting a legal precedent. The beneficiary of the book won the case, and the person responsible for putting the poems online was forced to withdraw the book from the internet and pay a consequent fine. The site would only allow to display one poem at a time [5]. Things could be worse today. Manifold approximation allows for a drastic reduction of the problem dimensionality, fundamentally modifying the infinite monkey theorem.\\

If Jonathan Basile and his library of babel [6] cannot reasonably claim ownership of every possible page of 3200 characters or less, what would it be if the same tool was not creating pages at random, but sampling from a relevant manifold? Applying this legal precedent to painting, would mean that the copyright applies to the generator and not only the outputs. It implies that one would be the owner of neural network's weights, which is quite a difficult concept to deal with.  In both cases, the ambiguity comes from our ability to free ourselves from the combinatorial absurdity (noise for an image, nonsense for a text) by using machine learning to approximate a text or image manifold that is close to the one we are able to conceptualize.

\section{The Question Concerning Art}

Is man just a tool in this process? This is a paradoxical question since according to Heiddger in \textit{The Question Concerning Technology} [7], art is precisely supposed to allow man to free himself from the challenging-forth revealing of modern technology. By considering nature as a raw material for technical operations, modern technology place humans themselves in standing-reserve.\\

On the contrary, Art is supposed to be a self-illuminating revealing that allows man to get to a primal truth. But is AI Art revealing still a form of Greek \textit{poeisis}? By sampling \textit{ad nauseam} from some approximation of the "world manifold", it reveals nature only by demanding that it provide aesthetics, exactly like modern technology demand that it supply energy that can be stored or extracted. In both cases, by considering the world as a standing-reserve, man ends up being nothing more than the orderer of the standing reserve, and risks becoming standing-reserve himself: merely a tool.

\section{Discussion and Conclusion}

This new artistic movement doesn't seem to be compatible with the current system and art market. The historical trend is to glorify the artist, while in AI Art he tends to disappear behind the process itself. Open source creates a unique situation where artists collaborate to push forward art and science. As desirable as it can be, it is problematic for plagiarism, authorship and artists' remuneration and credit. \\

Eventually, this movement forces us to acknowledge that all pieces of art exist, in a latent space, a Library of Babel, or simply a marble block. The artist does nothing but unveiling them. Queneau's work shows us that one can be an artist by playing dice. At the same time he introduces a founding demarcation: the artist is the one who provides the spectators with a dice and a generator. Is AI Art any different? Probably. Because of manifold approximation through optimization, the realm of possibilities explodes, and man becomes a tool part a wider creative process. Then, what can artists do to ensure that this GANstruction [8] remains a self-illuminating revealing ?

\end{document}